\documentclass[11pt,letterpaper]{article}
\usepackage{iftex}
\ifPDFTeX
  \usepackage[T1]{fontenc}
  \usepackage[utf8]{inputenc}
  \usepackage{lmodern}
\else
  \usepackage{fontspec}
\fi
\usepackage[margin=1in]{geometry}
\usepackage{graphicx}
\usepackage{textcomp}
\usepackage{xcolor}
\usepackage{bm}
\usepackage{mathtools}
\usepackage{amssymb}
\usepackage{booktabs}
\usepackage{url}
\usepackage{algpseudocode}
\usepackage{algorithm}
\usepackage{multirow}
\usepackage{makecell}
\usepackage{threeparttable}
\usepackage{wrapfig}
\usepackage{placeins}
\usepackage{caption}
\usepackage{float}
\usepackage[numbers,sort&compress]{natbib}
\usepackage[hidelinks]{hyperref}
\DeclareMathOperator*{\eventually}{\Diamond}
\DeclareMathOperator*{\always}{\Box}
\DeclareMathAlphabet{\mathmybb}{U}{bbold}{m}{n}

\title{SafeDiSTL: Decomposing Joint Constraints for Safe Multi-Agent Reinforcement Learning}

\author{Hongrui Zheng \quad Aditya Singh \quad Rahul Mangharam\\[0.5em]
  University of Pennsylvania, Philadelphia, USA\\
  \small\texttt{\{hongruiz,singhadi,rahulm\}@engineering.upenn.edu}}
\date{}

\begin{document}
\maketitle
\begin{abstract}
Cooperative multi-agent tasks often impose constraints on a shared objective or collective safety. A team level violation cost function can measure this constraint outcome, but does not distinguish which agents’ actions most strongly influence that violation. We present SafeDiSTL, a constrained multi-agent reinforcement learning framework that allows for credit assignment even with joint constraint measurements.
We use Signal Temporal Logic (STL) robustness to measure episode-level violation through running-maximum cost increments. Smooth full-episode robustness and differentiable dynamics provide action sensitivities, which allocate each cost increment across agents at the same decision time. These sensitivities account for delayed effects when weighting agents’ shares of each cost increment.
We study the framework in cooperative payload transport, connected navigation, and networked traffic control tasks.
SafeDiSTL achieves 1.6-5.1x the safe-success rate in transport, improves navigation safe-success by 11.1\%, and delivers 50-68\% more throughput in traffic when compared to constrained baselines. Results demonstrate a strong improvement of the learned policy in dealing with shared constraint while retaining the performance of unconstrained policies.
\end{abstract}

\begin{figure}[htbp]
    \centering
    \includegraphics[width=0.985\textwidth]{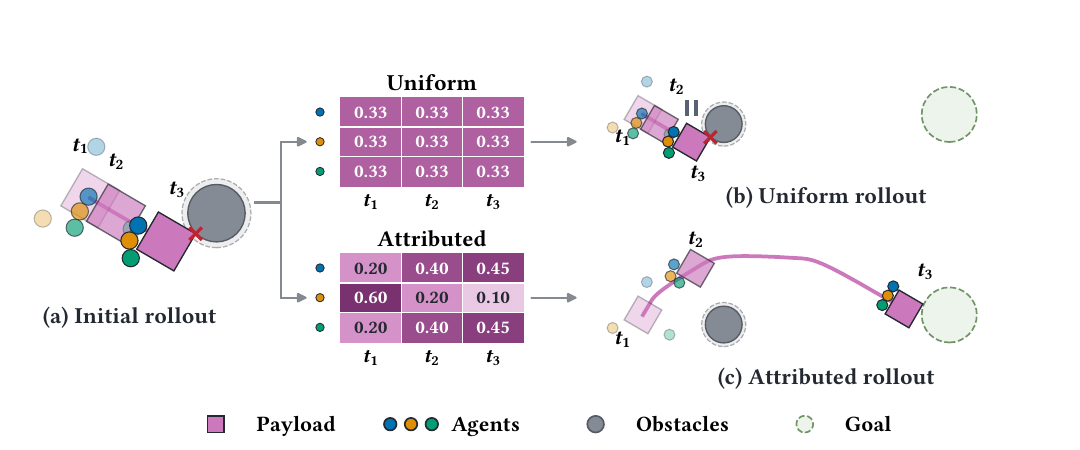}
    \caption{An illustration of the responsibility attribution mechanism. (a) A shared payload approaches an obstacle and incurs a constraint violation. (b) Uniform responsibility allocation produces conflicting corrections and agents stay deadlocked. (c) Agent-specific attribution enables coordinated obstacle avoidance.}
    \label{fig:overview}
\end{figure}

\section{Introduction}
Recent developments in multi-agent autonomous systems and their applications in aerial swarms, intelligent transportation networks, and automated warehouses have made multi-agent coordination a central problem.
In these environments, multiple agents must operate in shared spaces and achieve collective objectives, including collaborative manipulation, routing, and formation control, while satisfying system constraints such as connectivity and collision avoidance.
Constraints in single-agent RL are commonly formulated with Constrained Markov Decision Processes (CMDPs)~\cite{altman_cmdp}. Policies are then optimized by reformulating the problem as a primal-dual optimization between reward and costs. Constrained Policy Optimization (CPO)~\cite{achiam2017cpo} balances updates with more reward or cost consideration using a Lagrangian multiplier.
Multi-agent Reinforcement Learning (MARL) is a popular paradigm for learning decentralized agent policies in multi-agent problems. Existing work extends CMDP formulations to MARL~\cite{gu_safe_2023}, although generally assuming well-defined, decoupled cost functions for individual agents.
However, a shared constraint definition is important in multi-agent settings. For example, if multiple robots carry a shared rigid payload, a crash between the payload and obstacles is hard to attribute to individual agents.
Consider the example shown in Figure~\ref{fig:overview}: three warehouse robots collaboratively transport a large payload to a designated delivery location. At a given point $t_3$, one robot might be required to push in a direction such that the payload avoids the obstacle (green agent pushing up), while the other two maintain progress toward the goal (yellow and blue push right).
When encountering a violation such as that in Figure~\ref{fig:overview}.(a), assigning equal responsibility to all agents leads to a deadlock (Figure~\ref{fig:overview}.(b)) without addressing the asymmetrical desired updates to individual agents.
This shows that a shared constraint definition alone is not enough. It tells the team of agents how much they violated, but without explicit learning signals to individual agents on how to update their policies. Without careful decomposition, the introduction of a uniform penalty for all agents will hinder task learning progress. If an agent makes significant contribution towards the team's task progress but not towards violating constraint at a timestep, that individual transition should not incur a penalty in the policy learning process.

Credit assignment is an active area of research in MARL. Counterfactual Multi-Agent Policy Gradients (COMA)~\cite{foerster2018coma} established counterfactual reasoning for agents on task reward. Existing work~\cite{gu_safe_2023,okawa_multi-agent_2023,ling_constrained_2023} considers both individual and shared constraints in MARL problems, but leaves much to be desired in the effective decomposition of the constraint objective. This motivates the need for a principled mechanism to decompose the shared constraint among agents while accounting for the changing individual responsibilities to enable effective collaboration.
Signal Temporal Logic (STL)~\cite{maler_monitoring_2004} is a compact formal specification language that expresses logical and temporal requirements over trajectories. The quantitative measure called robustness~\cite{donze_robust_2010,fainekos_robustness_2008} assigns the margin of success to a trajectory instead of a binary signal.
The \emph{biggest advantage} of using STL is that differentiating its robustness through the full trajectory captures how individual actions influence subsequent constraint outcomes. We use these gradients to allocate cost increments across agents at its original timestep, while cost advantages propagate learning signals to earlier decisions.
For the example in Figure~\ref{fig:overview}, we express the shared responsibility as "Always no collision between payload and obstacle". Differentiable dynamics help achieve this by propagating individual agents' influence to joint state space trajectories, and allowing back propagation from the margin of violation to decisions. As shown in Figure~\ref{fig:overview}.(c), the green agent learns to subtly change its behavior by pushing \textit{upward} rather than to the right, resulting in a successful avoidance of the obstacle.
 In summary, the contributions of this work are as follows:
\begin{enumerate}
    \item We introduce an agent- and time-specific attribution mechanism that decomposes shared team constraints using STL robustness gradients while preserving the total cost.
    \item We develop SafeDiSTL, a constrained MARL framework that incorporates these attributed costs into policy optimization for decentralized execution.
    \item We evaluate SafeDiSTL on three tasks, comparing task performance and constraint satisfaction with other baselines.
\end{enumerate}



\section{Related Work}

\textit{Constrained MARL:}
Gu et al.~\cite{gu_safe_2023} introduce MACPO and MAPPO-Lagrangian, using constrained trust region updates and Lagrangian policy optimization, respectively.
Like MACPO, IsCPO~\cite{okawa_multi-agent_2023} coordinates separate agent policies through sequential updates. IsCPO considers global costs and updates agents in random order, passing optimization information to subsequent agents. Thus, later agents account for the effects of earlier updates on the shared constraint. Both MACPO and IsCPO coordinate policy updates, rather than an explicit allocation of a trajectory's violation cost according to each agent's influence.
Ling et al.~\cite{ling_constrained_2023} directly address constraint credit assignment through fictitious MARL in a collective setting. Its formulation exploits anonymous interactions represented by aggregate counts of agents in local states and actions, rather than individual identities. This structure supports scalable reward and constraint credit assignment, but is not assumed in our setting, where shared constraints can depend on coupled geometric configurations and action-dependent dynamics. SafeDiSTL instead obtains agent-specific cost allocations from the gradient sensitivity of a shared trajectory specification.

\textit{Multi-agent Credit Assignment:}
Credit assignment methods for multi-agent systems extract individual learning signals from shared team outcomes. COMA ~\cite{foerster2018coma} uses a centralized critic and a counterfactual baseline that marginalizes one agent's action while holding the others fixed. QPD~\cite{yang2020qpd} applies integrated gradients along trajectory paths to decompose global action values into individual credits. STAS~\cite{chen2023stas} first redistributes episodic return over time and then uses Shapley attribution between agents. These methods establish counterfactual, gradient-based, and temporal attribution for cooperative reward optimization. Our focus is the attribution of the shared constraint: we differentiate STL robustness through dynamics rather than attribute a learned return prediction.

\textit{Responsibility Allocation:}
Wang et al.~\cite{wang2025stlmarl} use STL robustness to construct multi-agent rewards incorporating task and safety specifications.
Zheng et al.~\cite{zheng_stl-svpio_2026} uses STL robustness as a learning signal for coordination of multiple agents.
Lyu et al.~\cite{lyu2023responsibility} dynamically distribute collision avoidance responsibility using a control barrier function (CBF)-inspired risk measure.
Cosner et al.~\cite{cosner2023responsibility} learn scene dependent allocations for responsibility aware control barrier functions.
These approaches demonstrate the value of asymmetric safety responsibility in decentralized control.
SafeDiSTL addresses the related problem of allocating learning signals for constraint violations during policy optimization, using STL robustness gradients to construct agent specific update signals rather than allocating online safety filter obligations.

\section{Preliminaries}
\label{sec:prelim}

\subsection{Signal Temporal Logic}
\label{sec:prelim_stl}
\textbf{Syntax:}
Signal Temporal Logic (STL)~\cite{maler_monitoring_2004} allows us to create spatiotemporal requirements over trajectories $\tau$, which encodes the global constraint in our learning problem.
For example, we can express ``Always avoid a collision between payload and obstacles'' in a formal specification: $\phi=\Box_{[0,5]}\neg\mu_{\mathrm{col}}(p_{\mathrm{payload}}, p_{\mathrm{obs}})$, where $\mu_{\mathrm{col}}$ is a predicate function that measures whether the two objects described by positions $p$ are in collision.
The STL syntax is
\begin{equation}\label{eq:stl_syntax}
  \phi := \top \mid \mu^g \mid \neg\phi \mid \phi_1\wedge \phi_2 \mid \phi_1\,\mathbf{U}_{[t_1,t_2]}\,\phi_2.
\end{equation}
$\top$ is the truth value, $\mu^g:=g(\tau(t))\ge 0$ is an atomic predicate, its truth value being determined by the value of the function $g:\mathbb{R}^n\rightarrow\mathbb{R}$.
$\wedge$ and $\neg$ denote the ``conjunction'' and ``negation'' Boolean operators, respectively.
$\mathbf{U}$ denotes the ``until'' temporal operator, which specifies that formula $\phi_1$ holds until $\phi_2$ holds in interval $[t_1,t_2]$.
Disjunction is defined as $\phi_1 \vee \phi_2 := \neg(\neg\phi_1 \wedge \neg\phi_2)$;
``Eventually'' is defined as $\eventually_{[t_1,t_2]}\phi := \top\,\mathbf{U}_{[t_1,t_2]}\,\phi$;
``Always'' is defined as $\always_{[t_1,t_2]}\phi := \neg\eventually_{[t_1,t_2]}\neg\phi$; 
And ``Implication'' is defined as $\phi_1\implies\phi_2:=\neg\phi_1\vee\phi_2$.

\textbf{Quantitative Semantics:}
STL uses \textbf{robustness} $\rho^\phi(\tau)\in\mathbb{R}$ to measure how strongly a given trajectory satisfies or violates a specification $\phi$. $\rho>0$ implies satisfaction and $\rho<0$ implies violation. We used the standard definition to calculate robustness~\cite{donze_robust_2010,fainekos_robustness_2008}.
We also utilize smooth approximations~\cite{pant_smooth_2017,leung2023stlcg, kapoor_stlcg_2025} of max and min operators to enable auto-differentiation of the robustness value of a trajectory to the states in the trajectory. The end-to-end gradient computation from robustness to policy actions can then be enabled with differentiable dynamics.

\section{Problem Formulation}
\subsection{Constrained Cooperative Markov Game}
\label{sec:prob_constrained_markov_game}
The cooperative multi-agent Markov Decision Process (MDP) is defined by the tuple:
\begin{equation*}
    \mathcal{M} = \bigl< \mathcal{S}, \mathcal{A}=\Pi_{i=1}^{N}\mathcal{A}_i, P, r, \rho_0, \gamma \bigr>,
\end{equation*}
where $\mathcal{S}$ is the state space, $\mathcal{A}$ is the joint action space, $\mathcal{A}_i$ is the action of agent $i$ out of $N$, $P$ is the transition kernel, $r$ is the shared task reward, $\rho_0$ is the initial state distribution, and $\gamma$ is the discount factor.
We extend this standard multi-agent MDP to a constrained Markov Game by extending the tuple to:
\begin{equation*}
    \mathcal{M}_C=\bigl<\mathcal{M}, c, d \bigr>,
\end{equation*}
where $c:\mathcal{S}\times\mathcal{A}\rightarrow\mathbb{R}_{\geq0}$ is a non-negative cost function representing the shared responsibility of all agents, and $d$ is a scalar cost budget for each episode.
In this setting, we consider decentralized agent policies, where the joint policy factorizes as follows:
\begin{equation*}
    \pi_{\theta}(\boldsymbol{a_t}|o_t) = \Pi_{i=1}^{N}\pi_{\theta}(a_{i,t}|o_{i,t}),
\end{equation*}
and the system evolves as:
\begin{equation*}
    s_{t+1}\sim P(\cdot|s_t, \boldsymbol{a_t}),
\end{equation*}
where $\theta$ is the shared policy parameters, $\boldsymbol{a_t}$ is the joint action produced at time $t$, $o_{i,t}$ is the observation of agent $i$ at time $t$, and $s_t$ is the joint state at time $t$. The shared task reward objective can be expressed as:
\begin{equation}
J_R(\boldsymbol{\theta})=\mathbb{E}_{\tau\sim\pi_{\theta}}\biggl[\sum_{t=0}^{H-1}\gamma^tr(s_t,\boldsymbol{a_t})\biggr],
\end{equation}
where $\tau\sim\pi_{\theta}$ denotes the trajectories realized under policy $\pi_\theta$, and $H$ denotes the episode horizon.
In addition to the task reward, we also consider the shared cost objective:
\begin{equation}
\label{eq:cost_obj}
    J_C(\boldsymbol{\theta})=\mathbb{E}_{\tau\sim\pi_{\theta}}\biggl[\sum_{t=0}^{H-1}\gamma^tc(s_t,\boldsymbol{a_t})\biggr].
\end{equation}
Hence, the combined constrained team objective of the agents becomes:
\begin{equation}
\label{eq:original_objective}
    \operatorname{maximize}_{\theta} ~J_R(\boldsymbol{\theta}),~~~~~\operatorname{subject~to} ~J_C(\boldsymbol{\theta}) \leq d
\end{equation}

\subsection{Lagrangian SafeMARL}
\label{sec:prob_safe_marl}
A standard Lagrangian formulation of the constrained objective in Equation~\eqref{eq:original_objective} introduces a nonnegative multiplier that trades reward against excess cost~\cite{achiam2017cpo}:
\begin{equation}
    \max_{\boldsymbol{\theta}}\min_{\lambda\geq0}\mathcal{L}(\boldsymbol{\theta}, \lambda)=J_R(\boldsymbol{\theta})-\lambda(J_C(\boldsymbol{\theta})-d)
\end{equation}
A canonical first-order projected primal--dual update performs ascent in the policy parameters and descent in the multiplier. At iteration $k$, it takes the form
\begin{equation}
\label{eq:lagrangian_updates}
\begin{aligned}
    \theta_{k+1} &= \theta_k+ \eta_{\theta}\nabla_{\theta} \mathcal{L}(\theta_k,\lambda_k),\\
    \lambda_{k+1} &=\bigl[\lambda_k+\eta_{\lambda}\bigl(J_C(\theta_k)-d\bigr)\bigr]_+,
\end{aligned}
\end{equation}
where $[\cdot]_+=\max(0,\cdot)$ and $\eta_{\theta},\eta_{\lambda}>0$ are learning rates. The multiplier increases when the cost exceeds the budget. Equation~\eqref{eq:lagrangian_updates} is an idealized gradient template: practical algorithms use sampled estimates and may change the surrogate objective, optimizer, or update schedule. Gu et. al.~\cite{gu_safe_2023} proposes two constrained MARL algorithms MACPO and MAPPO-L: MACPO solves sequential constrained trust-region subproblems rather than using the multiplier recursion in Equation~\ref{eq:lagrangian_updates}; MAPPO-L instead combines reward and cost advantages in a PPO-style surrogate with nonnegative multiplier updates. Our proposed method SafeDiSTL follows the Lagrangian approach but attributes the shared trajectory cost across agents and decision times for the actor update, while retaining the unallocated shared cost for multiplier updates.

\section{SafeDiSTL}
\label{sec:safedistl}
A shared constraint cost measures the team’s violation but does not distinguish the contributions of individual agents and decisions. SafeDiSTL uses full-episode STL sensitivities to allocate each running-maximum cost increment across agents at the same decision time. Section~\ref{sec:safedistl_cost} develops this allocation while preserving both the per-step increment and the total episode cost. Section~\ref{sec:safedistl_learning} incorporates the attributed costs into critic learning and constrained policy optimization, while retaining the original episode cost for the multiplier update.

\subsection{Decomposing the Shared Constraint}
\label{sec:safedistl_cost}
SafeDiSTL uses an STL specification on the joint state of all agents for both calculating the shared constraint cost and credit assignment. Let $\tau=(s_0,\boldsymbol a_0,\ldots,\boldsymbol a_{T_\tau-1},s_{T_\tau})$ be a complete episode. Let $H$ denote the maximum episode length and $T_\tau\leq H$ the number of transitions in episode $\tau$, which may terminate before reaching $H$. Let $\rho^\phi(\tau)$ denote the episode's robustness with respect to the shared constraint specification $\phi$. We define the episode violation cost and its expectation as
\begin{equation}
\label{eq:exact_stl_cost}
\begin{aligned}
C^\phi(\tau)&=[-\rho^\phi(\tau)]_+,\\
J_C^\phi(\boldsymbol\theta)&=\mathbb E_{\tau\sim\pi_\theta}[C^\phi(\tau)].
\end{aligned}
\end{equation}
In this paper, we consider the always-conjunction specifications type. This type of specification follows this form:
\begin{equation}
\label{eq:transport_stl_cost}
\begin{aligned}
\phi&=\always_{[0,T_\tau-1]}\bigwedge_{j=1}^{K}\mu^{g_j},\\
\rho^\phi(\tau)&=\min_{0\le t<T_\tau}\min_{1\le j\le K}g_j(s_{t+1}),
\end{aligned}
\end{equation}
where $K$ counts shared predicates and $g_j(s)\ge0$ indicates satisfaction of predicate $j$. Each predicate function $g(j)$ here is essentially a distance function that describe a part of the shared constraint. For example, it could be the Euclidean distance between a shared payload and an obstacle. Inspired by State-wise Constrained Policy Optimization~\cite{zhao_state-wise_nodate}, we accumulate the episode cost through increments in the running maximum. Define the instantaneous post-transition violation of the shared constraint and running maximum by
\begin{equation}
\label{eq:running_max_cost}
\begin{aligned}
c_{\mathrm{ins}}(s_{t+1})&=\left[-\min_{1\le j\le K}g_j(s_{t+1})\right]_+,\\
c_{\mathrm{max},t}&=\max\{c_{\mathrm{max},t-1},c_{\mathrm{ins}}(s_{t+1})\},
\end{aligned}
\end{equation}
with $c_{\mathrm{max},-1}=0$. The incremental maximum cost is
\begin{equation}
\label{eq:incremental_max_cost}
\begin{aligned}
\Delta c_t&=c_{\mathrm{max},t}-c_{\mathrm{max},t-1}\\
&=[c_{\mathrm{ins}}(s_{t+1})-c_{\mathrm{max},t-1}]_+.
\end{aligned}
\end{equation}
Only a violation exceeding the maximum of the previous episode incurs an additional cost. These increments telescope to the episode violation cost:
\begin{equation}
\label{eq:incremental_stl_identity}
\sum_{t=0}^{T_\tau-1}\Delta c_t=c_{\mathrm{max},T_\tau-1}=C^\phi(\tau).
\end{equation}
Unlike the generic discounted cost in Equation~\eqref{eq:cost_obj}, our shared constraint objective sums without discounting. This preserves the peak violation irrespective of when it occurs, and the correctness of Equation~\eqref{eq:incremental_stl_identity}.
The increments identify when the measured violation increases, but not which decisions influence it. To construct agent- and time-specific learning signals, we differentiate trajectory robustness through the dynamics. Let $\widetilde\rho^\phi$ denote the differentiable smooth approximation of the robustness and define the action sensitivity
\begin{equation}
\label{eq:action_sensitivity}
z_{i,t}=
\left\|\frac{\partial[-\widetilde\rho^\phi(\tau)]}
{\partial a_{i,t}}\right\|_2.
\end{equation}
Note that we use the 2-norm here, thus our action sensitivity only measures the magnitude of the influence of an agent.
The derivative propagates through the entire recorded episode, holding the other sampled actions fixed. 
Thus, the sensitivities capture delayed effects rather than only immediate effects. The smooth robustness approximation and attribution normalization are detailed in Appendix~\ref{sec:stl_replay_details}.
We normalize these sensitivities across agents separately at each decision time:
\begin{equation}
\label{eq:attribution_weights}
w_{i,t}=\frac{z_{i,t}}{\displaystyle\sum_{j=1}^{N}z_{j,t}},
\end{equation}
assuming nonzero sensitivity at time $t$, with uniform weights when all sensitivities vanish. Thus, $\sum_iw_{i,t}=1$ for every valid decision time. We allocate the corresponding cost increment using detached attribution weights:
\begin{equation}
\label{eq:attributed_cost}
\widetilde c_{i,t}=N\Delta c_t\operatorname{stopgrad}(w_{i,t}).
\end{equation}
The sensitivity that determines the relative agent weights is from the full episode, while the costs allocated remain local at each timestep. If $\Delta c_t=0$, no attributed cost is assigned at that time. The factor $N$ accounts for averaging across agents, preserving each increment and hence the episode cost:
\begin{equation}
\label{eq:attribution_conservation}
\begin{aligned}
\frac{1}{N}\sum_{i=1}^{N}\widetilde c_{i,t}&=\Delta c_t,\\
\frac{1}{N}\sum_{i=1}^{N}\sum_{t=0}^{T_\tau-1}\widetilde c_{i,t}
&=\sum_{t=0}^{T_\tau-1}\Delta c_t=C^\phi(\tau).
\end{aligned}
\end{equation}

\subsection{Constrained Policy Optimization}
\label{sec:safedistl_learning}
We optimize decentralized policies using a MAPPO-style~\cite{yu_surprising_nodate} proximal policy optimization (PPO) algorithm under centralized training. Following the Lagrangian formulation in Section~\ref{sec:prob_safe_marl}, the policy update balances the shared task reward with the constraint cost. SafeDiSTL replaces the shared cost signal in this update with the agent-specific attributed costs defined in Section~\ref{sec:safedistl_cost}.

A centralized reward critic $V_R(s_t)$ estimates the expected discounted share task return. For constraint costs, we use an agent-specific critic $V_C^i(o_{i,t},c_{\mathrm{max},t-1})$ and a centralized critic $V_C^g(s_t,c_{\mathrm{max},t-1})$. They estimate the expected remaining cumulative cost assigned to agent $i$ and its team mean, respectively, using the allocation from Section~\ref{sec:safedistl_cost}. The corresponding cost signals are $\widetilde c_{i,t}$ and
\begin{equation}
\label{eq:mean_attributed_cost}
\overline c_t=\frac{1}{N}\sum_{i=1}^{N}\widetilde c_{i,t}
=\Delta c_t.
\end{equation}
Attribution changes the distribution of cost across agents but not its timing or team mean. The centralized cost critic therefore learns from the original increments $\Delta c_t$, while the agent-specific critic learns from $\widetilde c_{i,t}$. Both receive $c_{\mathrm{max},t-1}$ as training-only context for the constraint history.
Then generalized advantage estimation (GAE) connects each signal and critic to its advantage:
\begin{equation}
\label{eq:safedistl_gae_connection}
A_{R,t}=\operatorname{GAE}_t(r,V_R),\quad
A^C_{i,t}=\operatorname{GAE}_t(\widetilde c_i,V_C^i),\quad
A^C_{g,t}=\operatorname{GAE}_t(\overline c,V_C^g).
\end{equation}
Here, GAE takes the corresponding signal sequence and rollout-time critic predictions. The critic regression targets are:
\begin{equation}
\label{eq:safedistl_critic_targets}
y_{R,t}=A_{R,t}+V_{R,t},\qquad
y^C_{i,t}=A^C_{i,t}+V^i_{C,t},\qquad
y^C_{g,t}=A^C_{g,t}+V^g_{C,t}.
\end{equation}
Here $V_{R,t}$, $V^i_{C,t}$, and $V^g_{C,t}$ denote predictions during rollout.
The centralized cost critic supplies a joint-state reference for the local critics: a consistency penalty encourages their mean prediction to agree with it. Therefore, it influences the actor indirectly through agent-specific predictions, rather than through a separate $A^C_{g,t}$ term in the actor objective.
We combine the shared reward advantage with each agent's attributed cost advantage:
\begin{equation}
\label{eq:safedistl_actor_advantage}
\begin{aligned}
A_{i,t}
&=A_{R,t}-\lambda_kA^C_{i,t},\\
q_{i,t}(\theta)&=\frac{\pi_\theta(a_{i,t}\mid o_{i,t})}
{\pi_{\theta_k}(a_{i,t}\mid o_{i,t})}.
\end{aligned}
\end{equation}
Here $\lambda_k\ge0$ is the shared constraint multiplier at training iteration $k$, and $q_{i,t}(\theta)$ is the action likelihood ratio between the updated and rollout policies. The actor is updated using the PPO clipped objective
\begin{equation}
\label{eq:safedistl_policy_objective}
\begin{aligned}
\mathcal L_{\mathrm{actor}}(\theta)
&=\mathbb{E}_{i,t}\bigl[\min\{
q_{i,t}(\theta)A_{i,t},\\
&\qquad\operatorname{clip}(q_{i,t}(\theta),1-\epsilon,1+\epsilon)
A_{i,t}\}\bigr],
\end{aligned}
\end{equation}
where $\epsilon>0$ is the clipping threshold. The advantages and attribution weights are held fixed during this update. When individual constraints are also present, their cost advantages and multipliers contribute separate penalty terms and are not included in the shared attribution.
For the dual update in Equation~\eqref{eq:lagrangian_updates}, we estimate $J_C^\phi(\theta_k)$ using the mean exact episode cost $C^\phi(\tau)$, not the attributed cost of any single agent. The shared multiplier $\lambda_k$ increases when this mean exceeds $d$ and decreases when it falls below $d$, subject to its nonnegative range. Thus, attribution changes the agents' policy-learning signals while the dual update uses the original shared constraint.
Algorithm~\ref{alg:safedistl} summarizes the training procedure. Cost-advantage estimation, critic losses, and implementation details are given in Appendix~\ref{sec:safedistl_optimization_details}.

\begin{algorithm}[t]
\caption{SafeDiSTL}
\label{alg:safedistl}
\begin{algorithmic}[1]
\State Initialize $\pi_\theta$, critics $V_R,V_C^i,V_C^g$, and $\lambda_0$.
\For{each training iteration $k$}
  \State Collect complete episodes using $\pi_{\theta_k}$.
  \State Compute $C^\phi(\tau)$, increments $\Delta c_t$, and full-episode sensitivities $z_{i,t}$. 
  \State Normalize $z_{i,t}$ across agents at each $t$ to obtain $w_{i,t}$.
  \State Allocate $\widetilde c_{i,t}=N\Delta c_t\operatorname{stopgrad}(w_{i,t})$; set $\overline c_t=\Delta c_t$.
  \State Compute $A_{R,t},A^C_{i,t},A^C_{g,t}$ using Equation~\eqref{eq:safedistl_gae_connection}.
  \State Fit $V_R,V_C^i,V_C^g$ to $y_{R,t},y^C_{i,t},y^C_{g,t}$, with cost consistency regularization.
  \State Form $A_{i,t}=A_{R,t}-\lambda_k A^C_{i,t}$.
  \State Update $\theta_k$ to $\theta_{k+1}$ using Equation~\eqref{eq:safedistl_policy_objective}.
  \State Update $\lambda_k$ to $\lambda_{k+1}$ from the mean $C^\phi(\tau)$ and budget $d$.
\EndFor
\end{algorithmic}
\end{algorithm}

\section{Experiments}
\label{sec:experiments}

We evaluate SafeDiSTL on three cooperative multi-agent benchmarks with different shared constraints: package--obstacle clearance in transport, communication connectivity in navigation, and core occupancy in traffic control. Across tasks, we examine whether attribution of a shared constraint supports task completion while maintaining constraint satisfaction. We compare against reward-only MAPPO and the constrained baselines MACPO and MAPPO-L from Gu et. al.~\cite{gu_safe_2023}. Baseline optimization details are provided in Appendix~\ref{sec:baseline_details}.

\subsection{Experimental Setup}
\label{sec:experimental_protocol}

\begin{figure*}[!t]
\centering
\includegraphics[width=\textwidth]{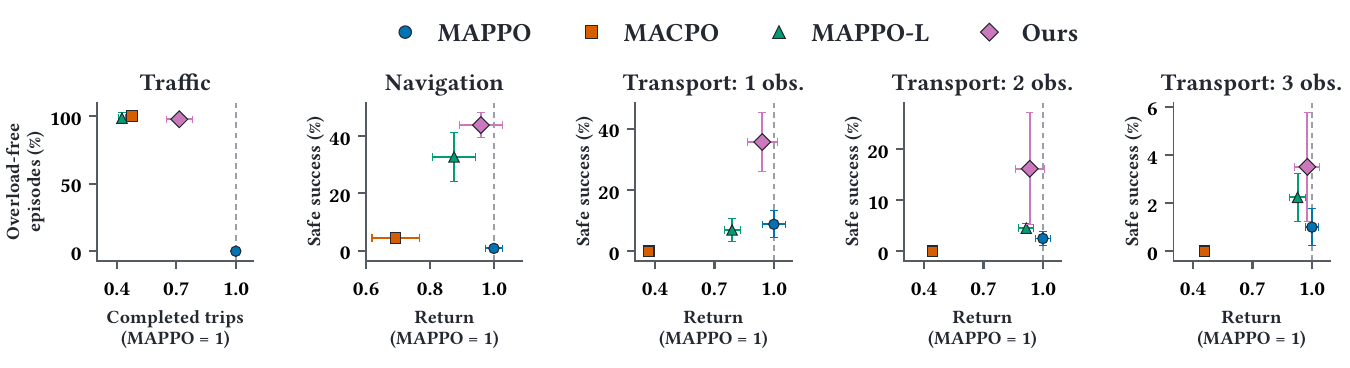}
\caption{Constraint--performance tradeoffs across benchmarks. Points and error bars show means and sample standard deviations across six training seeds. Traffic reports overload-free episode percentages versus completed trips. Others report safe-success rates where goal was reached without constraint violation versus task return. $x$-axes metrics are normalized by MAPPO values. Note that the sub-plots zooms into parts of the $y$-range for better visibility and comparison, and hence have different $y$-axes ranges.}
\label{fig:benchmark_tradeoff}
\end{figure*}

\begin{figure*}[t]
\centering
\includegraphics[width=\textwidth]{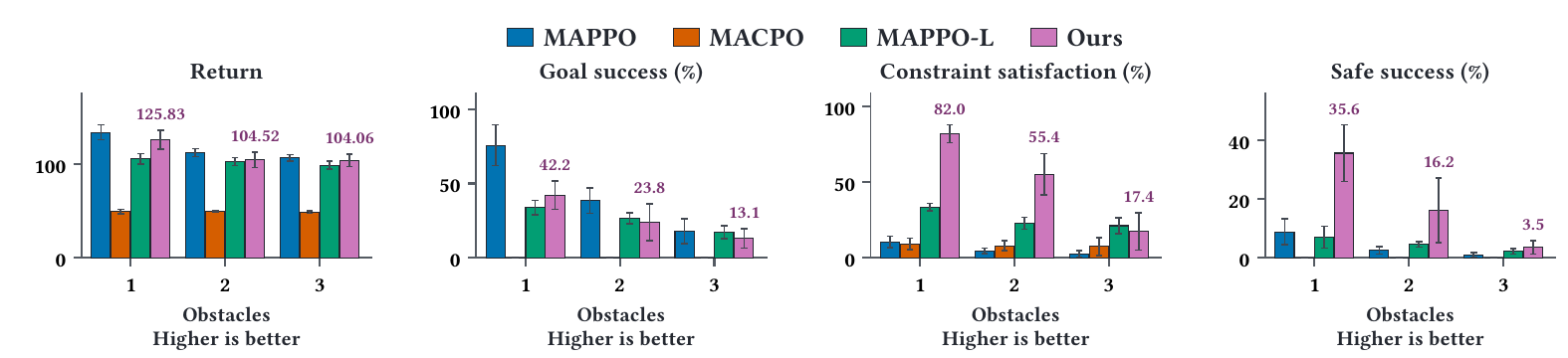}
\caption{Transport results across obstacle counts (mean $\pm$ SD over six training seeds). Labels are SafeDiSTL mean values.}
\label{fig:transport_results}
\end{figure*}

All methods use six training seeds and common evaluation resets within each benchmark. We compare task success, shared constraint satisfaction, and their joint occurrence, as well as task returns. All methods are trained with the same compute budget in each scenario in terms of simulation steps.
Transport policies train with 15M sim steps, and evaluates on 500 common resets per obstacle condition. Navigation policie train with 15M sim steps, and evaluates on 128 common resets. Traffic policies train with 20M sim steps and evaluates on 1,000 common resets.
In all benchmarks, SafeDiSTL uses the running max increment cost defined in Secion~\ref{sec:safedistl_cost}. All other benchmarks utilizes cost functions equivalent to the per step cost of the same STL specification. For example, in transport, SafeDiSTL's STL specification says "Always no collision between the payload and the obstacles", the equivalent per step cost function is the overlapping margin in Euclidean distance between the payload and obstacles' geometric shapes. Each method then pre-processes the cost functions according to the reference implementation. All experiments are conducted in JAX~\cite{jax2018github} on GPUs.  We primarily examine four key metrics. 
\begin{enumerate}
    \item Task return: episodic sum of task reward that the policies achieve in evaluation.
    \item Goal success rate: the percentage of episodes that leads to successful goal reaching.
    \item Constraint satisfaction rate: the percentage of evaluation episodes that did not incur any constraint violations.
    \item Safe success rate: the percentage of episodes that has the joint occurrence of goal success and constraint satisfaction.
\end{enumerate}

Environment details and policy architectures are provided in Appendices~\ref{sec:experiment_details} and~\ref{sec:observation_policy_details}, respectively.

\begin{figure}[t]
\centering
\includegraphics[width=0.85\linewidth]{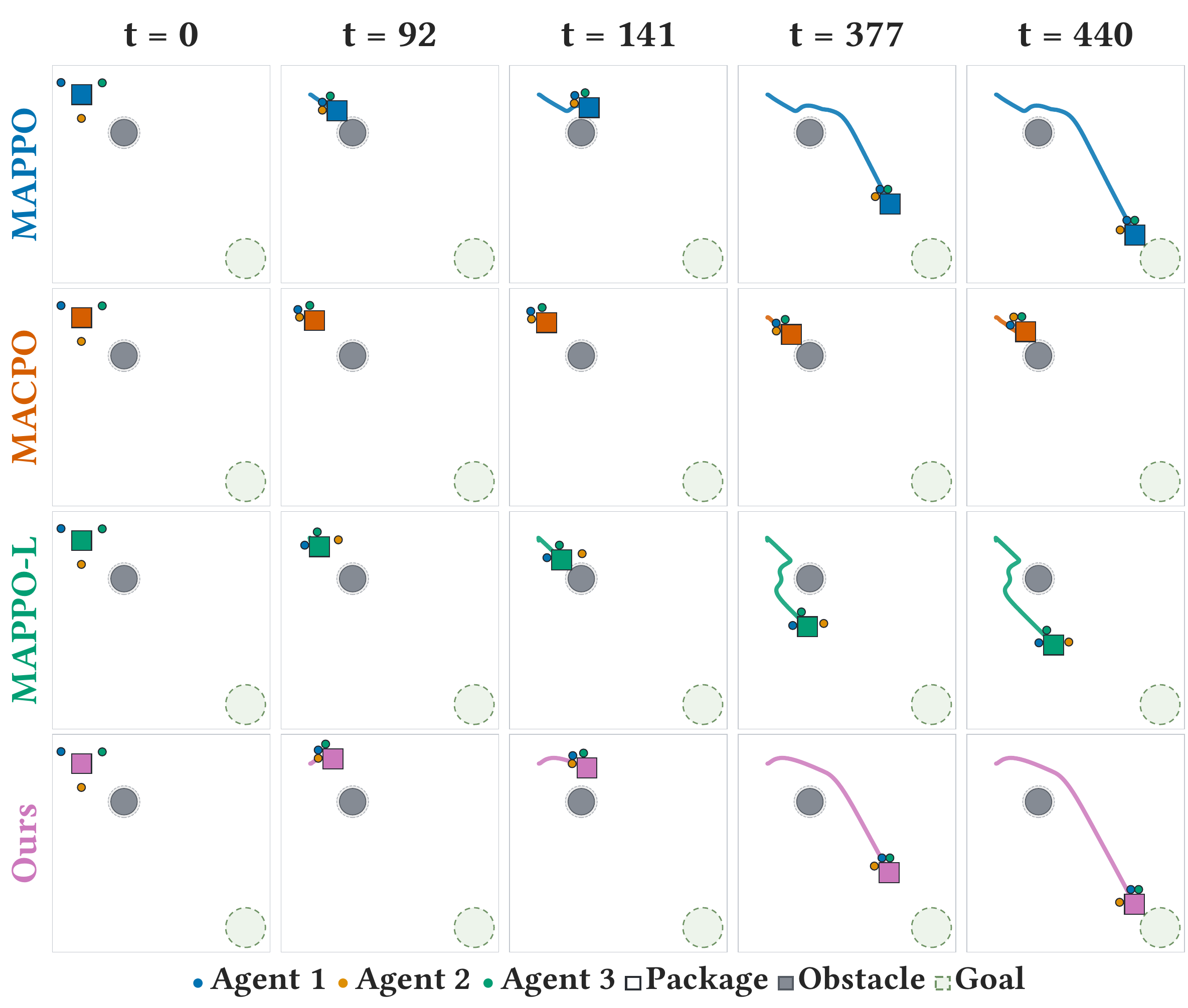}
\caption{Example rollout comparison with baselines in the 1-obstacle transport scenario with common initial conditions.}
\label{fig:transport_keyframes}
\end{figure}

\subsection{The Constraint-performance Tradeoff}
We first show the key result of the experiments. In Figure~\ref{fig:benchmark_tradeoff}, we show that SafeDiSTL balances the tradeoff between constraint satisfaction and retaining performance. In each subplot, the $y$-axes shows constraint satisfaction metrics in percentages, and the $x$-axes shows performance metrics in normalized returns to those obtained by MAPPO. Our algorithm modifies MAPPO by introducing the dual on constraint and asymmetric cost attribution to individual agents, thus we want to show that SafeDiSTL retains the performance of MAPPO, but does so considering satisfying the shared constraints. In all five scenarios shown in Figure~\ref{fig:benchmark_tradeoff}, SafeDiSTL balances this trade off by being in the top right quadrant of the plots, while other methods sacrifice MAPPO's task performance to increase constraint satisfaction. Appendix~\ref{sec:ablations} examines attribution, cost formulation, and dual-optimizer variants.

\subsection{Cooperative Transport}
\label{sec:exp_transport_2d}
We use a transport task with dynamics and control similar to the one in~\cite{bettini2022vmas}, where three force input agents push a shared package around circular obstacles towards a goal. The shared constraint here is package--obstacle clearance, and safe success requires both the payload reaching the goal and no contact between the payload and the obstacle throughout the entire episode. In this benchmark, we consider three different settings with 1-3 obstacles between the starting position and the payload goal.
Figure~\ref{fig:transport_results} shows the comparison between the four methods. SafeDiSTL has the highest mean safe success rate in all three settings: $35.6\pm9.6\%$, $16.2\pm11.0\%$, and $3.5\pm2.3\%$ with one, two, and three obstacles, respectively. MAPPO-L trails closest behind with $7.0\pm3.7\%$, $4.5\pm0.9\%$, and $2.2\pm1.0\%$. With three obstacles, safe success remains low for all methods in this harder condition, yet SafeDiSTL still achieves highest safe success rate there. Although MAPPO has higher mean goal success in all three conditions, SafeDiSTL more often achieves the goal while satisfying the shared clearance requirement. MACPO on the other hand completely fails in all three settings. Full numerical results appear in Table~\ref{tab:transport_2d}.
Figure~\ref{fig:transport_keyframes} shows an example rollout comparison between the methods. The same trend shows up here as well: MAPPO reaches the goal, the two constrained baselines has degraded goal reaching capability, while SafeDiSTL manages to go around the obstacle and reaches the goal.

\subsection{Connected Navigation}
\label{sec:exp_navigation}

\begin{figure}[t]
\centering
\includegraphics[width=0.85\linewidth]{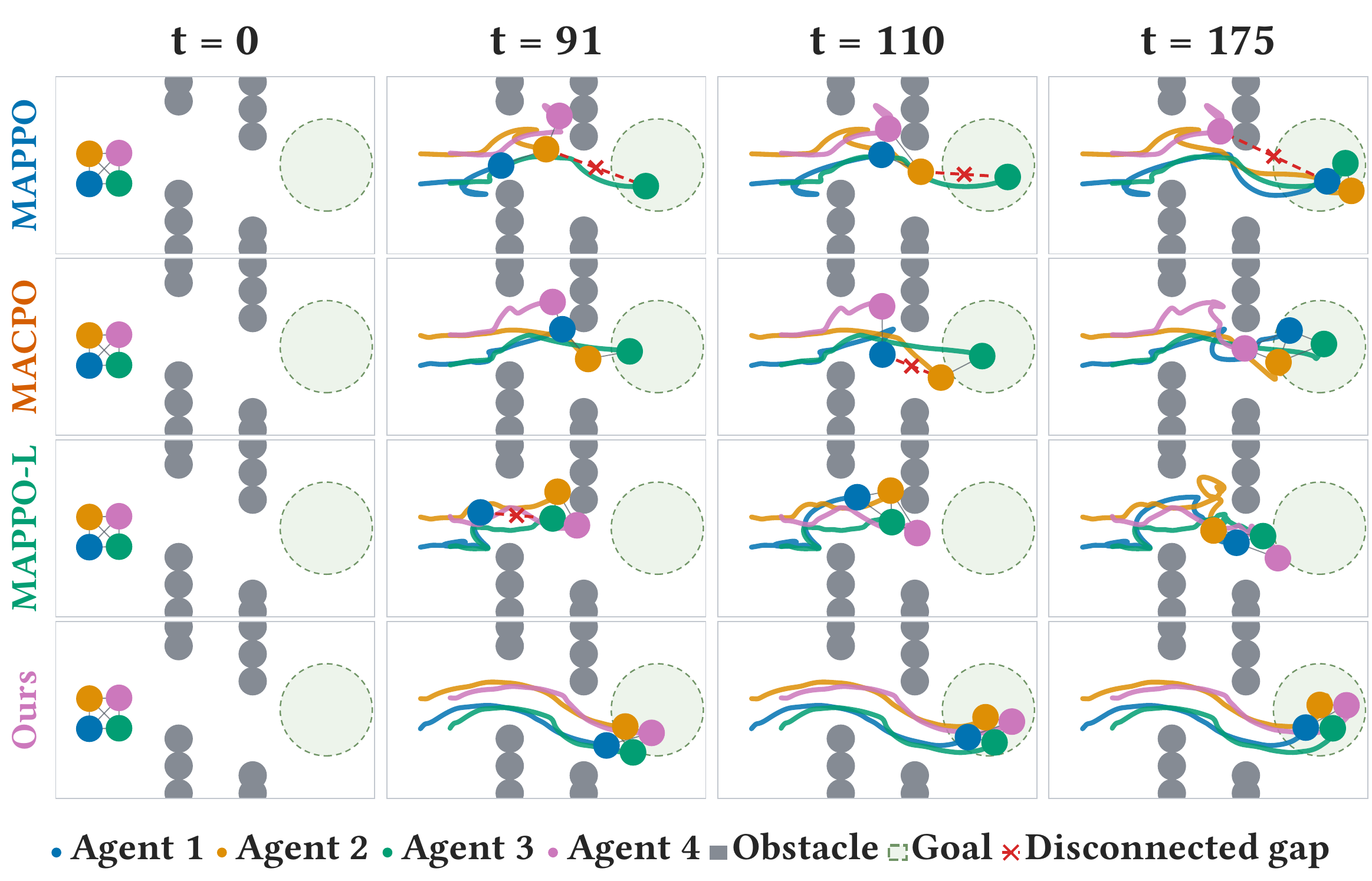}
\caption{Example rollout comparison with baselines in the connected navigation benchmark with common initial conditions.}
\label{fig:navigation_keyframes}
\end{figure}

\begin{figure}[t]
\centering
\includegraphics[width=\linewidth]{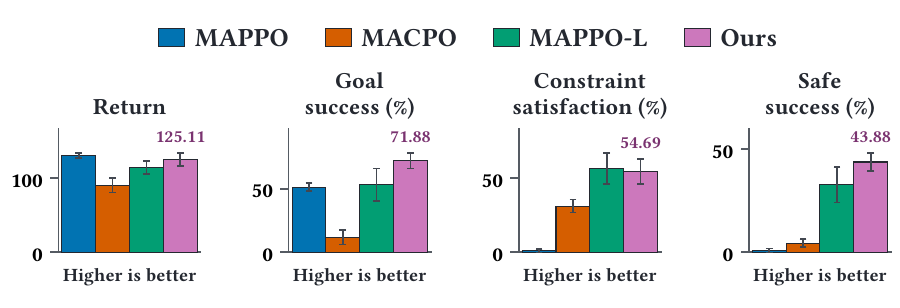}
\caption{Navigation results (mean $\pm$ SD over six training seeds). Labels are SafeDiSTL mean values.}
\label{fig:navigation_results}
\end{figure}

In this benchmark, we control four point-mass agents navigating through two staggered obstacle passages to a shared destination. The shared constraint is buffered full episode communication connectivity: at every timestep, the longest edge of the agents' Euclidean minimum-spanning tree must not exceed $3.325$ m. Goal success requires all agents to remain in the destination region during the final 20 control steps.
Figure~\ref{fig:navigation_results} reports the comparison between all four methods.
SafeDiSTL has the highest mean goal success rate ($71.88\pm6.13\%$) and safe-success rate ($43.88\pm4.40\%$) among the four main methods. MAPPO-L attains slightly higher full episode constraint satisfaction rate than SafeDiSTL ($56.64\pm10.61\%$ versus $54.69\pm8.33\%$). MACPO reaches only $4.43\pm1.96\%$ safe successes. MAPPO has the highest mean return but almost always violates the constraint. These results show that in the connected navigation benchmark, considering the constraint effectively also helps with task success rate since grouping together through the narrow passage reduces accidental collision with the obstacle that might slow down goal reaching. Full numerical results appear in Table~\ref{tab:navigation}.
Figure~\ref{fig:navigation_keyframes} also shows the same conclusion in matched rollout trajectories. SafeDiSTL agents grouped closely together and navigated smoothly towards the goal without any disconnects in the graph. The agents in all other methods disconnected at least once throughout the episode.

\subsection{Networked Traffic Control}
\label{sec:exp_traffic}

\begin{figure}[h]
\centering
\includegraphics[width=0.6\linewidth]{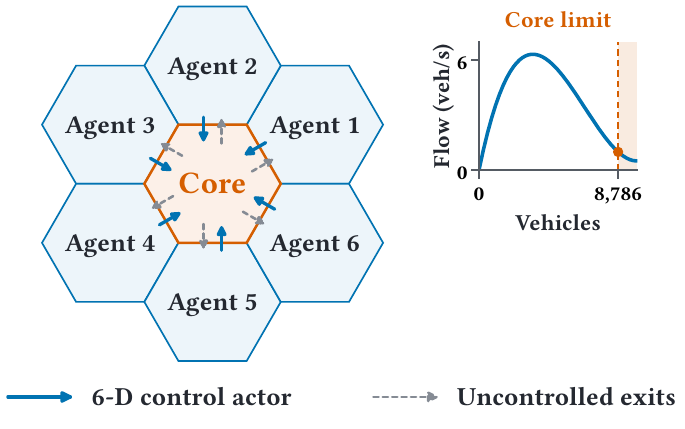}
\caption{Seven region traffic network with a shared core occupancy constraint. Controls for each agent is a 6-D vector that controls the flow rate for all 6 destinations (0 always for self).}
\label{fig:traffic_task}
\end{figure}

We consider a seven region Networked Macroscopic Fundamental Diagram (NMFD) traffic task~\cite{godfrey1969mechanism,tumu_differentiable_2024}. Six peripheral regions exchange destination labeled traffic through a shared central core. Each region meters its outgoing traffic, while core exits are uncontrolled and randomized at every time step. The shared constraint is on the core region's occupancy. The flow rate of the core region is dictated by the cubic polynomial shown in Figure~\ref{fig:traffic_task}. The flow rate first increases as the number of vehicles in the region increases. After the peak, it drops down to the congestion limit, where the flow rate drops below 1 vehicle per second. Thus the shared constraint is set as $n_{\mathrm{core}}\leq n_{\mathrm{safe}}\simeq8786.415$. Vehicle admissions to the core therefore trade off completed trips against core region congestion, while restrictive metering can increase the wait time in the system.

\begin{figure}[htbp]
\centering
\includegraphics[width=\linewidth]{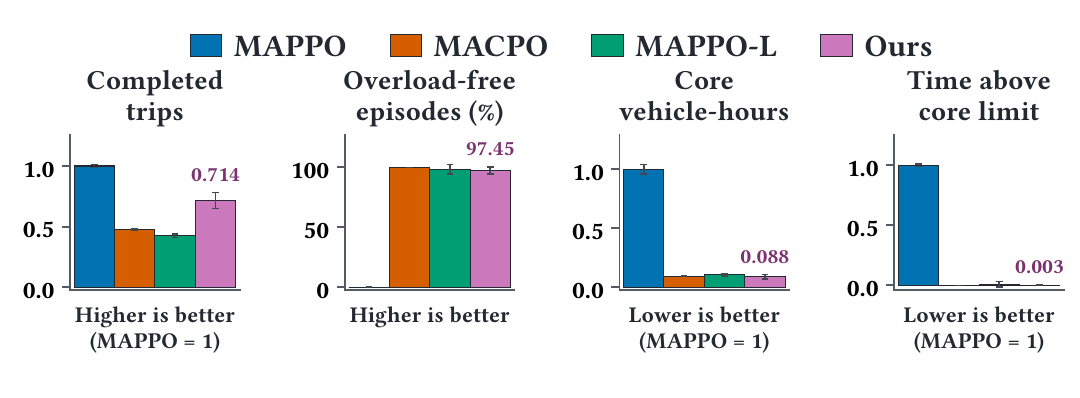}
\caption{Traffic results (mean $\pm$ SD over six training seeds). Labels are SafeDiSTL mean values.}
\label{fig:traffic_results}
\end{figure}
The traffic in the system is randomly spawned as source-destination pairing of two peripheral zones into the queue. All vehicles must travel through the core on their way to their destinations. The agents control their own regions' flow rate for vehicles with all other peripheral regions as destination. Then the system revolves following the controlled core region in-flow rate, and the uncontrolled, randomized out-flow rate. The agents observe the randomize exit rate, and should react to changes for decisions on their control.
For this task, since there isn't a clear measure for success, we use these key metrics instead:
\begin{enumerate}
    \item Completed trips: this is the equivalent to task return, where the policy uses as training reward.
    \item Overload-free episode rate: the percentage of full episodes without overloading the core region.
    \item Core vehicle-hours: the time that vehicles spends on average at the core region, this measures the system delay.
    \item Time above core limit: the time that core regions has occupancy higher than the limit.
\end{enumerate}
Figure~\ref{fig:traffic_results} compares all four methods. The training reward for all methods here are completed trips. The goal of this benchmark is to test the trade off between throughput and core congestion. Since core region exit rates are randomized, a smart team of agents should learn to coordinate when to let each destination's vehicles go without overloading the core.
SafeDiSTL retains $71.4\%$ of MAPPO throughput while reducing the core vehicle-hours to $8.8\%$ of MAPPO and the time above core limits to $0.3\%$ of MAPPO. Although MACPO and MAPPO-L also learn to reduce constraint violation to the same level, they only retain $47.6\%$ and $42.5\%$ of the MAPPO throughput, respectively. Full numerical results appear in Table~\ref{tab:traffic}.

\section{Limitations and Conclusions}
We propose SafeDiSTL, a constrained MARL framework that allows for credit assignment under shared team constraints. Using STL robustness gradients, we derive a cost attribution scheme that decomposes the shared team constraints while preserving the total cost. These attributed costs are then incorporated in policy optimization for centralized training and decentralized execution. Across three diverse case studies on cooperative payload transport, connected navigation, and traffic network control, SafeDiSTL shows a strong improvement in handling shared constraints over the other state-of-the art MARL baselines.

First, while SafeDiSTL has shown an exceptional safety performance tradeoff in benchmarks, it relies on the gradients of the robustness signal through the system dynamics. The dynamics gradients might be hard to come by in contact-rich scenarios. Even with non-differentiable dynamics, we can still estimate via finite differences by using more calls to the simulation. Future extensions should aim to mitigate this issue by devising gradient-free signals for STL robustness decomposition.
Second, our sensitivity metric used to allocate costs is the magnitude of influence. Future work could also consider how to incorporate the direction of influence. 
Third, we did not show benchmarks on higher fidelity robotic scenarios, and potential transfer of learned policies to real robots. Future works will include validation of the proposed method on real-world robotic platforms.

\bibliographystyle{ACM-Reference-Format} 
\bibliography{references}

\clearpage
\appendix
\section{Baseline Optimization Details}
\label{sec:baseline_details}
This appendix gives the optimization details summarized in Section~\ref{sec:prob_safe_marl}. Equation~\ref{eq:lagrangian_updates} is a first-order Lagrangian template, not the update rule of every constrained policy optimizer.
CPO~\cite{achiam2017cpo} distinguishes this stateful multiplier update from its constrained trust-region subproblem, whose multipliers are solved anew at each iteration. SafeMARL extends constrained policy optimization to cooperative Markov games through MACPO and MAPPO-Lagrangian~\cite{gu_safe_2023}.

\subsection{MACPO}
Multi-Agent Constrained Policy Optimisation (MACPO) updates decentralized actors sequentially, accounting for the policies of agents already updated in the current iteration~\cite{gu_safe_2023}. Each actor maximizes a reward surrogate subject to cost constraints and a Kullback--Leibler (KL) trust region. Thus, coordination enters the sequential surrogate rather than a simultaneous update of all actors using a fixed cost penalty.

In the shared-cost notation of Equation~\ref{eq:original_objective}, the local CPO-style approximation for agent $i$ has the form~\cite{achiam2017cpo}
\begin{equation}
\label{eq:macpo_local_step}
\begin{aligned}
\max_{\Delta\theta_i}\quad
&g_i^\top\Delta\theta_i,\\
\text{subject to}\quad
&c_i+b_i^\top\Delta\theta_i\leq0,\\
&\tfrac12\Delta\theta_i^\top H_i\Delta\theta_i\leq\delta_i.
\end{aligned}
\end{equation}
Here $g_i$ and $b_i$ are gradients of the sequential reward and shared-cost surrogates, $c_i$ is the current surrogate cost minus $d$, $H_i$ is the local Hessian of the mean KL divergence, and $\delta_i>0$ is the trust-region radius. These quantities are evaluated at the current actor parameters, with preceding actors' updates accounted for in the surrogates. Linearizing reward and cost while quadratically approximating KL yields a constrained natural-gradient step. For invertible $H_i$ and a positive trust-region multiplier $\eta_i$, its direction satisfies
\begin{equation}
\label{eq:macpo_dual_direction}
\begin{aligned}
\Delta\theta_i^{\star}
&=\eta_i^{-1}H_i^{-1}(g_i-\nu_i b_i),\\
\theta_i^{\mathrm{new}}
&=\theta_i^{\mathrm{old}}+\alpha_i\Delta\theta_i^{\star},
\end{aligned}
\end{equation}
where $\nu_i\geq0$ enforces the local cost constraint and $\alpha_i\in(0,1]$ is selected by backtracking. Unlike the stateful $\lambda_k$ in Equation~\ref{eq:lagrangian_updates}, these subproblem multipliers are solved anew for each actor update.

MACPO uses a cost-reducing recovery step when the constrained update is infeasible~\cite{gu_safe_2023}. Multiple budgets add linearized cost constraints to Equation~\ref{eq:macpo_local_step}. The theoretical sequential procedure provides reward-improvement and cost-feasibility bounds under its assumptions; sampling error, local approximations, and finite line searches mean that practical neural-policy updates do not guarantee constraint satisfaction on every trajectory.

\subsection{MAPPO-Style Lagrangian Surrogates}
A first-order Lagrangian implementation replaces the exact policy gradient in Equation~\ref{eq:lagrangian_updates} with a sample-based surrogate using reward and cost advantages. Let $A_{R,t}$ and $A_{C,t}$ be reward and shared-cost advantage estimates obtained from centralized critics. For the decentralized actor of agent $i$, define
\begin{equation}
\label{eq:lagrangian_advantage_ratio}
\begin{aligned}
A^{\lambda}_{i,t}
&=A_{R,t}-\lambda_kA_{C,t},\\
q_{i,t}(\theta_i)
&=\frac{\pi_{\theta_i}(a_{i,t}\mid o_{i,t})}
{\pi_{\theta_{i,k}}(a_{i,t}\mid o_{i,t})}.
\end{aligned}
\end{equation}
A MAPPO-style clipped surrogate for a fixed multiplier is
\begin{equation}
\label{eq:lagrangian_ppo}
\begin{aligned}
\widehat{\mathcal L}_{\mathrm{clip}}(\boldsymbol{\theta})
&=\widehat{\mathbb E}_{i,t}\bigl[
\min\{q_{i,t}(\theta_i)A^{\lambda}_{i,t},\\
&\qquad\operatorname{clip}(q_{i,t}(\theta_i),1-\epsilon,1+\epsilon)
A^{\lambda}_{i,t}\}\bigr],
\end{aligned}
\end{equation}
where $\epsilon>0$ is the clipping threshold and the empirical expectation is over valid on-policy samples. Centralized critics may access $s_t$, while execution uses only $o_{i,t}$. Homogeneous agents can share actor parameters without changing the factorized action sampling. Multiple constraints introduce one nonnegative multiplier per budget and a sum of cost-advantage terms.

The surrogate above illustrates the reward--cost combination rather than specifying every detail of MAPPO-L. The Algorithm-3 MAPPO-L baseline used in our experiments updates independent actors sequentially and resets its nonnegative multipliers at each actor's optimization phase; its cost objectives and budgets are specified with each benchmark. Reward-only MAPPO omits the cost term. SafeDiSTL uses attributed cost advantages instead of the broadcast shared-cost advantage, as described in Section~\ref{sec:safedistl}.

Clipping and a finite number of stochastic primal--dual updates do not enforce exact feasibility at each iteration. In particular, this first-order implementation does not inherit CPO's trust-region bounds or the guarantees of SafeMARL's theoretical sequential update. Moreover, broadcasting $A_{C,t}$ supplies no explicit differentiation of constraint responsibility across agents. Section~\ref{sec:safedistl} replaces this broadcast signal with trajectory-dependent attribution while retaining the exact shared constraint for dual accounting.

\section{Supplementary Transport Results}
\label{sec:transport_results_table}
Table~\ref{tab:transport_2d} provides the numerical results plotted in Figure~\ref{fig:transport_results}.

\begin{table*}[t]
\centering
\small
\caption{Supplementary Transport results compared to baselines. Entries are mean $\pm$ sample SD across six seeds and 500 common resets per seed. Rates are percentages. Safe success denotes joint occurrence of goal reaching and shared-constraint satisfaction.}
\label{tab:transport_2d}
\begin{tabular}{clrrrr}
\toprule
Obstacles & Algorithm & Return & Goal success & \makecell{Constraint\\ satisfaction} & Safe success\\
\midrule
1 & MAPPO & $133.68\pm7.96$ & $75.7\pm13.7$ & $10.4\pm3.8$ & $8.8\pm4.4$\\
1 & MACPO & $49.25\pm2.23$ & $0.0\pm0.0$ & $9.1\pm3.9$ & $0.0\pm0.0$\\
1 & MAPPO-L & $105.49\pm5.40$ & $33.9\pm4.8$ & $33.4\pm2.5$ & $7.0\pm3.7$\\
1 & Ours & $125.83\pm10.24$ & $42.2\pm9.7$ & $82.0\pm5.9$ & $35.6\pm9.6$\\
\midrule
2 & MAPPO & $111.98\pm4.19$ & $38.6\pm8.6$ & $4.5\pm1.9$ & $2.5\pm1.3$\\
2 & MACPO & $49.65\pm0.91$ & $0.0\pm0.0$ & $8.0\pm3.4$ & $0.0\pm0.0$\\
2 & MAPPO-L & $102.56\pm4.15$ & $26.6\pm3.6$ & $22.8\pm4.0$ & $4.5\pm0.9$\\
2 & Ours & $104.52\pm8.15$ & $23.8\pm12.5$ & $55.4\pm13.6$ & $16.2\pm11.0$\\
\midrule
3 & MAPPO & $106.49\pm3.50$ & $17.9\pm8.3$ & $2.6\pm2.2$ & $1.0\pm0.8$\\
3 & MACPO & $48.69\pm1.27$ & $0.0\pm0.0$ & $7.5\pm5.9$ & $0.0\pm0.0$\\
3 & MAPPO-L & $98.79\pm4.35$ & $17.1\pm4.3$ & $21.2\pm5.1$ & $2.2\pm1.0$\\
3 & Ours & $104.06\pm6.74$ & $13.1\pm6.6$ & $17.4\pm12.4$ & $3.5\pm2.3$\\
\bottomrule
\end{tabular}
\end{table*}

\section{Supplementary Navigation Results}
\label{sec:navigation_results_table}
Table~\ref{tab:navigation} provides the numerical results plotted in Figure~\ref{fig:navigation_results}, together with safe return and final-20 distance.
\begin{table*}[t]
\centering
\small
\setlength{\tabcolsep}{4pt}
\caption{Supplementary Navigation results compared to baselines. Entries are mean $\pm$ sample SD across six seeds and 128 common resets per seed. Goal success, constraint satisfaction, and safe success are percentages; safe success requires both goal achievement and whole-episode connectivity. Safe return assigns zero to unsafe episodes; final 20 step distance is in meters.}
\label{tab:navigation}
\begin{tabular}{lrrrrrr}
\toprule
Algorithm & Return & \makecell{Goal success\\(\%)} & \makecell{Constraint\\satisfaction (\%)} & \makecell{Safe success\\(\%)} & Safe return & \makecell{Final 20 Step\\distance}\\
\midrule
MAPPO & $130.43\pm3.38$ & $51.17\pm3.31$ & $0.91\pm1.04$ & $0.78\pm0.99$ & $1.49\pm1.72$ & $2.14\pm0.16$\\
MACPO & $90.32\pm9.70$ & $11.85\pm5.79$ & $31.25\pm4.39$ & $4.43\pm1.96$ & $29.24\pm3.29$ & $2.59\pm0.23$\\
MAPPO-L & $114.14\pm8.71$ & $52.99\pm12.67$ & $56.64\pm10.61$ & $32.81\pm8.62$ & $69.80\pm10.19$ & $1.84\pm0.27$\\
Ours & $125.11\pm8.80$ & $71.88\pm6.13$ & $54.69\pm8.33$ & $43.88\pm4.40$ & $75.31\pm10.50$ & $1.83\pm0.30$\\
\bottomrule
\end{tabular}%
\end{table*}

\section{Supplementary Traffic Results}
\label{sec:traffic_results_table}
Table~\ref{tab:traffic} retains the full traffic metrics, complementing Figure~\ref{fig:traffic_results}.

\begin{table*}[t]
\centering
\normalsize
\setlength{\tabcolsep}{5pt}
\caption{Supplementary Traffic results relative to MAPPO (column-wise MAPPO mean $=1$). Mean $\pm$ SD across six training seeds; 1,000 common reset scenarios per policy. Higher completed trips and lower values in all other columns are preferable.}
\label{tab:traffic}
\resizebox{\linewidth}{!}{%
\begin{tabular}{lrrrrr}
\toprule
Method & Completed trips & \makecell{Network\\vehicle-hours} & \makecell{Core\\vehicle-hours} & \makecell{Episode-max\\core cost} & \makecell{Time above\\core limit}\\
\midrule
MAPPO & $1.000\pm0.012$ & $1.000\pm0.001$ & $1.000\pm0.041$ & $1.000\pm0.081$ & $1.000\pm0.010$\\
MACPO & $0.476\pm0.007$ & $1.053\pm0.001$ & $0.091\pm0.002$ & $0.000\pm0.000$ & $0.000\pm0.000$\\
MAPPO-L & $0.425\pm0.016$ & $1.058\pm0.002$ & $0.104\pm0.007$ & $0.00074\pm0.00180$ & $0.009\pm0.023$\\
Ours & $0.714\pm0.066$ & $1.029\pm0.007$ & $0.088\pm0.018$ & $0.000061\pm0.000053$ & $0.003\pm0.004$\\
\bottomrule
\end{tabular}
}
\end{table*}

\section{Ablation Studies}
\label{sec:ablations}

We ablate responsibility attribution, the cost formulation, and the dual optimizer. The four variants use uniform attribution with either max-incremental cost or step cost, with fixed-step or Adam multiplier updates. Max-incremental cost is the same incremental maximum cost $\Delta c_t$ defined for SafeDiSTL in Equation~\ref{eq:incremental_max_cost}; step cost uses positive per-transition violation. Labels without ``+ Adam'' use fixed-step multiplier updates. Ours retains sensitivity-based attribution with max-incremental cost. The Ours rows are repeated from the main tables, with the same evaluation protocols and seed-level aggregation. Results of ablations are presented in Tables~\ref{tab:transport_2d_ablation}, \ref{tab:navigation_ablation} and \ref{tab:traffic_ablation}.

\begin{table*}[t]
\centering
\small
\caption{Two-dimensional transport ablations for one through three obstacles. Rates are percentages; entries are mean $\pm$ sample SD across six training seeds, with 500 common resets per seed. All three Ours rows are copied unchanged from Table~\ref{tab:transport_2d}.}
\label{tab:transport_2d_ablation}
\resizebox{\linewidth}{!}{%
\begin{tabular}{clrrrr}
\toprule
Obstacles & Algorithm & Return & Goal success & \makecell{Shared\\constraint sat.} & Safe success\\
\midrule
1 & \makecell[l]{Uniform attribution\\ Max-incremental cost} & $58.91\pm8.30$ & $0.5\pm0.8$ & $74.4\pm6.3$ & $0.2\pm0.4$\\
1 & \makecell[l]{Uniform attribution\\ Max-incremental cost + Adam} & $70.00\pm26.21$ & $8.4\pm8.7$ & $75.3\pm6.5$ & $5.6\pm7.7$\\
1 & \makecell[l]{Uniform attribution\\ Step cost} & $46.81\pm10.14$ & $0.0\pm0.0$ & $57.7\pm9.1$ & $0.0\pm0.0$\\
1 & \makecell[l]{Uniform attribution\\ Step cost + Adam} & $44.26\pm5.41$ & $0.0\pm0.0$ & $58.6\pm11.7$ & $0.0\pm0.0$\\
1 & Ours & $125.83\pm10.24$ & $42.2\pm9.7$ & $82.0\pm5.9$ & $35.6\pm9.6$\\
\midrule
2 & \makecell[l]{Uniform attribution\\ Max-incremental cost} & $59.40\pm6.96$ & $0.7\pm1.4$ & $61.8\pm8.9$ & $0.2\pm0.3$\\
2 & \makecell[l]{Uniform attribution\\ Max-incremental cost + Adam} & $80.63\pm23.60$ & $6.5\pm5.8$ & $73.3\pm10.1$ & $4.1\pm4.0$\\
2 & \makecell[l]{Uniform attribution\\ Step cost} & $38.41\pm3.55$ & $0.0\pm0.0$ & $60.1\pm18.6$ & $0.0\pm0.0$\\
2 & \makecell[l]{Uniform attribution\\ Step cost + Adam} & $44.65\pm7.49$ & $0.0\pm0.0$ & $52.1\pm13.1$ & $0.0\pm0.0$\\
2 & Ours & $104.52\pm8.15$ & $23.8\pm12.5$ & $55.4\pm13.6$ & $16.2\pm11.0$\\
\midrule
3 & \makecell[l]{Uniform attribution\\ Max-incremental cost} & $52.54\pm10.11$ & $0.0\pm0.0$ & $62.2\pm4.1$ & $0.0\pm0.0$\\
3 & \makecell[l]{Uniform attribution\\ Max-incremental cost + Adam} & $81.79\pm12.46$ & $1.7\pm1.1$ & $70.8\pm6.4$ & $0.6\pm0.7$\\
3 & \makecell[l]{Uniform attribution\\ Step cost} & $42.93\pm3.33$ & $0.0\pm0.0$ & $50.1\pm15.6$ & $0.0\pm0.0$\\
3 & \makecell[l]{Uniform attribution\\ Step cost + Adam} & $44.66\pm6.93$ & $0.0\pm0.0$ & $54.6\pm15.6$ & $0.0\pm0.0$\\
3 & Ours & $104.06\pm6.74$ & $13.1\pm6.6$ & $17.4\pm12.4$ & $3.5\pm2.3$\\
\bottomrule
\end{tabular}
}
\end{table*}

\begin{table*}[t]
\centering
\small
\setlength{\tabcolsep}{4pt}
\caption{Navigation ablations. Metrics and evaluation resets match Table~\ref{tab:navigation}; Ours is copied unchanged. Entries are mean $\pm$ sample SD across six training seeds. Goal, constraint-satisfaction, and safe-goal rates are percentages; final-20 distance is in meters.}
\label{tab:navigation_ablation}
\resizebox{\linewidth}{!}{%
\begin{tabular}{lrrrrrr}
\toprule
Algorithm & Return & Goal (\%) & Safe (\%) & Safe goal (\%) & Safe return & Final-20 distance\\
\midrule
\makecell[l]{Uniform attribution\\ Max-incremental cost} & $126.77\pm9.41$ & $54.30\pm15.19$ & $2.60\pm2.55$ & $2.60\pm2.55$ & $4.21\pm4.19$ & $2.13\pm0.34$\\
\makecell[l]{Uniform attribution\\ Max-incremental cost + Adam} & $128.58\pm5.08$ & $50.52\pm13.01$ & $0.39\pm0.65$ & $0.39\pm0.65$ & $0.65\pm1.10$ & $2.17\pm0.29$\\
\makecell[l]{Uniform attribution\\ Step cost} & $39.60\pm22.40$ & $12.63\pm9.34$ & $90.36\pm4.57$ & $11.98\pm9.11$ & $36.42\pm20.95$ & $5.29\pm1.66$\\
\makecell[l]{Uniform attribution\\ Step cost + Adam} & $129.38\pm15.43$ & $61.33\pm20.97$ & $11.98\pm8.81$ & $10.16\pm9.40$ & $19.28\pm14.75$ & $1.94\pm0.43$\\
Ours & $125.11\pm8.80$ & $71.88\pm6.13$ & $54.69\pm8.33$ & $43.88\pm4.40$ & $75.31\pm10.50$ & $1.83\pm0.30$\\
\bottomrule
\end{tabular}
}
\end{table*}

\begin{table*}[t]
\centering
\normalsize
\setlength{\tabcolsep}{5pt}
\caption{Traffic ablations. Values use the same fixed, column-wise MAPPO mean as Table~\ref{tab:traffic}; Ours is copied unchanged. Mean $\pm$ sample SD across six training seeds and 1,000 common resets per policy. Higher completed trips and lower values in all other columns are preferable.}
\label{tab:traffic_ablation}
\resizebox{\linewidth}{!}{%
\begin{tabular}{lrrrrr}
\toprule
Method & Completed trips & \makecell{Network\\vehicle-hours} & \makecell{Core\\vehicle-hours} & \makecell{Episode-max\\core cost} & \makecell{Time above\\core limit}\\
\midrule
\makecell[l]{Uniform attribution\\ Max-incremental cost} & $0.786\pm0.018$ & $1.021\pm0.002$ & $0.079\pm0.017$ & $0.00031\pm0.00058$ & $0.005\pm0.009$\\
\makecell[l]{Uniform attribution\\ Max-incremental cost + Adam} & $1.002\pm0.010$ & $1.000\pm0.001$ & $1.011\pm0.076$ & $1.002\pm0.111$ & $1.000\pm0.010$\\
\makecell[l]{Uniform attribution\\ Step cost} & $0.388\pm0.235$ & $1.063\pm0.025$ & $0.065\pm0.038$ & $0.00000049\pm0.00000076$ & $0.000041\pm0.000063$\\
\makecell[l]{Uniform attribution\\ Step cost + Adam} & $0.881\pm0.033$ & $1.011\pm0.003$ & $0.837\pm0.050$ & $0.788\pm0.082$ & $0.985\pm0.012$\\
Ours & $0.714\pm0.066$ & $1.029\pm0.007$ & $0.088\pm0.018$ & $0.000061\pm0.000053$ & $0.003\pm0.004$\\
\bottomrule
\end{tabular}
}
\end{table*}

\section{Robustness Differentiation}
\label{sec:stl_replay_details}
We write the differentiable dynamics as $s_{t+1}=F(s_t,\boldsymbol a_t)$, with sampled disturbances held fixed. The norm in Equation~\eqref{eq:action_sensitivity} is over the action coordinates of agent $i$. Differentiating negative robustness rather than its rectified violation avoids an additional zero-gradient region when the smoothed trajectory is safe; the allocated cost is nevertheless zero whenever the exact $C^\phi(\tau)$ is zero.

\paragraph{Attribution normalization in implementation.}
\label{sec:attribution_implementation}
Let $\mathcal I_t$ contain the active agents at valid decision time $t$, and let $Z_t=\sum_{i\in\mathcal I_t}z_{i,t}$. The sum in Equation~\eqref{eq:attribution_weights} excludes inactive entries. With numerical threshold $\varepsilon_g>0$, the masked normalization is
\begin{equation}
\label{eq:masked_attribution_weights}
w_{i,t}=\begin{cases}
z_{i,t}/Z_t,&i\in\mathcal I_t,\ Z_t>\varepsilon_g,\\
1/|\mathcal I_t|,&i\in\mathcal I_t,\ Z_t\leq\varepsilon_g,\\
0,&i\notin\mathcal I_t.
\end{cases}
\end{equation}
The uniform fallback handles finite, uninformative sensitivities and preserves $\sum_iw_{i,t}=1$ at each valid time with at least one active agent. Invalid padding receives zero weight; allocation assumes a nonempty active set at every cost-bearing transition. The factor $N$ in Equation~\eqref{eq:attributed_cost} corresponds to the fixed team-size mean, including zero contributions from inactive agents. Nonfinite active derivatives are flagged rather than assigned uniform credit. Complete episodes are required before full-episode sensitivities and attribution targets can be finalized, even though the cost allocation is time-local. Sensitivity norms discard derivative direction, and attribution depends on the model, action parameterization, and smoothing scale. Conservation of scalar cost does not make the attributed policy gradient an unbiased estimator of the original shared-cost gradient.

The exact minimum can concentrate its derivative on a single predicate and time. For the always-conjunction in Equation~\eqref{eq:transport_stl_cost}, we use the log-sum-exp approximation
\begin{equation}
\label{eq:smooth_stl_surrogate}
\widetilde\rho^\phi_\beta(\tau)
=-\frac{1}{\beta}\log\!\sum_{t=0}^{T_\tau-1}\sum_{j=1}^{K}
\exp\!\bigl(-\beta g_j(s_{t+1})\bigr),
\end{equation}
where $\beta>0$ controls smoothing. Other supported Boolean and temporal compositions use their corresponding smooth minimum or maximum. The surrogate supplies derivatives only: its value and sign do not replace the exact cost or the exact satisfaction test.

\section{SafeDiSTL Optimization and Implementation Details}
\label{sec:safedistl_optimization_details}
\subsection{Cost advantages and critic losses}
The hybrid cost critic consists of a centralized predictor $V_C^g(s_t,c_{\mathrm{max},t-1})$ and a shared local-observation predictor $V_C^i(o_{i,t},c_{\mathrm{max},t-1})$, evaluated separately for each agent. The local heads predict expected remaining cumulative attributed costs from $\widetilde c_{i,t}$ and the global head predicts their team mean from $\overline c_t$, defined in Equation~\eqref{eq:mean_attributed_cost}, with the numerical scale below. Their learned outputs are not an exact decomposition by architecture.

The regression targets in Equation~\eqref{eq:safedistl_critic_targets} use unnormalized GAE estimates and rollout-time value predictions, and are held fixed during critic fitting. Cost prediction and GAE use the same fixed numerical scale $\kappa_C>0$ on both $\widetilde c_{i,t}$ and $\overline c_t$. Let $b_t$ be one if the episode continues after transition $t$, and zero at either termination or the finite-horizon endpoint. With cost discount one, the agent-specific temporal-difference residual and generalized advantage estimate are
\begin{equation}
\label{eq:attributed_cost_gae}
\begin{aligned}
\delta^C_{i,t}
&=\kappa_C\widetilde c_{i,t}
+b_t V_C^i(o_{i,t+1},c_{\mathrm{max},t})
-V_C^i(o_{i,t},c_{\mathrm{max},t-1}),\\
A^C_{i,t}
&=\sum_{\ell=0}^{T_\tau-1-t}\eta_C^\ell\delta^C_{i,t+\ell},
\end{aligned}
\end{equation}
where $\eta_C\in[0,1]$ is the cost trace-decay parameter. The global cost estimate uses the same recursion with $\overline c_t$ and $V_C^g$. Cost targets do not bootstrap beyond the completed episode. The GAE-based cost targets $y^C_{i,t}$ and $y^C_{g,t}$ are defined in Equation~\eqref{eq:safedistl_critic_targets}, using fixed rollout-time value predictions. The cost critic is trained with clipped squared-error value losses and a consistency penalty:
\begin{equation}
\label{eq:hybrid_cost_loss}
\begin{aligned}
\mathcal L_C
&=\widehat{\mathbb E}_t\bigl[
\ell_v(V_C^g,y^C_{g,t})
+\tfrac1N\sum_i\ell_v(V_C^i,y^C_{i,t})\\
&\qquad+\zeta\bigl(\tfrac1N\sum_iV_C^i-V_C^g\bigr)^2\bigr].
\end{aligned}
\end{equation}
Here $\ell_v$ clips changes relative to the rollout-time value prediction, and $\zeta\geq0$ controls consistency. This regularizer encourages agreement of the predicted mean with the global prediction; Equation~\ref{eq:attribution_conservation} holds for the allocated costs independently of critic accuracy, not necessarily for the learned values or generalized advantage targets.

\subsection{Actor normalization and individual constraints}
If agent $i$ also has local constraints indexed by $j$, let $A^{\mathrm{loc}}_{C,i,j,t}$ and $\lambda^{\mathrm{loc}}_{i,j}$ denote their separately estimated, scaled cost advantages and multipliers. We form
\begin{equation}
\label{eq:safedistl_implementation_advantage}
A^{\mathrm{norm}}_{i,t}
=\mathcal N\!\left(
A_{R,t}-\lambda_kA^C_{i,t}
-\sum_j\lambda^{\mathrm{loc}}_{i,j}
A^{\mathrm{loc}}_{C,i,j,t}\right),
\end{equation}
where $\mathcal N$ denotes normalization of the combined advantage over valid actor samples. The implementation substitutes this normalized advantage into Equation~\eqref{eq:safedistl_policy_objective} and adds an entropy bonus during optimization. Individual obstacle and workspace costs do not enter the shared robustness gradient or its normalization. They retain their own critics, budgets, and dual updates. Thus, attributing a payload violation cannot erase an individual robot's constraint.

For a local clearance $\ell_{i,j,t}$ and required buffer $\Delta_j$, the local signal is
\begin{equation}
\label{eq:local_clearance_cost}
c^{\mathrm{loc}}_{i,j,t}=[\Delta_j-\ell_{i,j,t}]_+.
\end{equation}
Each local dual uses the mean raw deficit of its own agent and channel over active transitions, not an average across teammates. These are stepwise budgets, distinct from the shared episode-maximum budget. The local critic uses its configured discounted cost return and bootstraps clean horizon truncations, whereas the completed-episode shared STL return does not. In the three-dimensional suites, both local buffers are 0.05 m, local cost discount and trace decay are 0.99 and 0.95, and local multipliers use projected budget-relative updates with stepsize $10^{-3}$ and cap two.

\subsection{Shared multiplier update}
The shared multiplier is updated from the mean exact, unallocated cost of completed episodes. In the budget-relative, fixed-step configuration, the update is
\begin{equation}
\label{eq:safedistl_dual}
\begin{aligned}
\widehat C_k&=\frac{1}{B_k}\sum_{b=1}^{B_k}C^\phi(\tau_b),\\
\lambda_{k+1}&=\operatorname{clip}\!\left(
\lambda_k+\eta_\lambda\frac{\widehat C_k-d}{d},
0,\lambda_{\max}\right),
\end{aligned}
\end{equation}
with positive $d$ and $B_k$ completed episodes in the dual-update window. Neither $\kappa_C$ nor the attribution weights change this physical-unit constraint statistic. The finite multiplier cap is a practical optimization choice, not an exact solution of the unrestricted saddle-point problem.

\section{Supplementary Experimental Details}
\label{sec:experiment_details}

\subsection{Collaborative Transport}
\label{sec:transport_experiment_details}

The transport environment contains three force-controlled agents, an unactuated nonrotating square package, and circular obstacles in a planar workspace.  The package has side length $0.15$ m and mass $50$; each agent has mass one and radius $0.03$ m, and each obstacle has radius $0.10$ m.  Each agent selects a normalized two-dimensional action in $[-1,1]^2$, scaled by $0.6$ to produce the applied force.  The differentiable VMAS-style simulator uses float64 smooth contact, and four $0.025$ s physics substeps per control step.  Episodes last 500 control steps.

Resets sample one of eight layouts, apply a random quarter-turn and reflection, jitter each obstacle by at most $0.015$ m per coordinate, and randomize the agents' phase around the package.  The layouts obstruct the direct route while providing detours.

The shared constraint margin is the package-box signed distance to each obstacle, minus the obstacle radius and the required $0.02$ m clearance.  Goal achievement uses the task's box-distance threshold of approximately $0.1567$ m rather than a point-payload center-distance test.  The shared reward combines distance-progress shaping scaled by 100 with the campaign's approach and joint-push coordination terms, held fixed across algorithms.  Package contact is logged separately; it neither adds a reward penalty nor terminates the episode.

\subsection{Connected Navigation}
\label{sec:navigation_experiment_details}

Four homogeneous planar agents navigate from a compact start formation to a shared circular destination through two staggered obstacle passages.  The canonical corridor is 18 m long and 10 m wide; ten static circular obstacles form two five-obstacle barriers.  Agent and obstacle radii are 0.75 m and 0.8 m, respectively.  Resets randomize barrier locations, opening centers, and opening widths (3.4--3.8 m), perturb the start formation and destination, and apply a global rotation, reflection, and translation.  Agents use bounded two-dimensional acceleration commands, maximum speed 1.5 m/s, smooth contact forces, five physics substeps per 0.1 s control step, and a 250-step horizon. The shared reward combines reduction in mean agent-to-goal distance, scaled by $5/R$, with the fraction of agents inside the destination.  Goal success requires all four agents to remain within the 2.6 m goal radius throughout the final 20 control steps.

The formal constraint is whole-episode buffered graph connectivity.  Let $b_u$ be the longest edge in the Euclidean minimum-spanning tree of the agent positions at checked instant $u$.  For transition $t$, let $\mathcal U_t$ contain its source and five physics-substep endpoints.  With communication radius $R=3.5$ m and buffer fraction $\xi=0.05$,
\begin{equation}
\label{eq:navigation_connectivity_margin}
\begin{aligned}
b_u&=\max_{(i,k)\in\operatorname{MST}(x_u)}
      \|x_{i,u}-x_{k,u}\|_2,\\
m_t&=\min_{u\in\mathcal U_t}
      \frac{R(1-\xi)-b_u}{\xi R}.
\end{aligned}
\end{equation}
The buffered check requires $m_t\geq0$ for every transition, equivalently a checked bottleneck no larger than $R(1-\xi)=3.325$ m.  The 0.175 m buffer exceeds the maximum relative displacement of 0.06 m between substep checks, satisfying the bounded-speed certificate condition.  A safe episode passes every buffered check; a safe goal additionally satisfies the goal criterion. Smooth contacts do not count as connectivity violations, and separate obstacle and boundary cost channels are disabled.  Disconnection does not terminate an episode.

\subsection{Networked Traffic Control}
\label{sec:traffic_experiment_details}

The traffic task is a seven-region NMFD network with six peripheral regions and a shared core.  Direct peripheral-to-peripheral transfers are disabled.  Each region uses the cubic production function
\begin{equation}
G(n)=4.133\times10^{-11}n^3-8.282\times10^{-7}n^2+0.0042n,
\end{equation}
where $n$ is occupancy in vehicles and $G$ is in vehicles/s.  Episodes contain 240 control intervals of 30 s, each integrated with ten RK4 substeps.  Demand is five times the reference setting: 0.625 vehicles/s per nonlocal peripheral origin--destination pair, with $\pm20\%$ reset randomization.  Three destinations receive low exit baselines of 0.1 and three receive high baselines of 0.9, randomly permuted at reset; temporal noise and incidents vary core exits within $[0,1]$.

Agent $i$ controls a six-dimensional admission vector $u_i\in[0,1]^6$. Coordinate $u_{id}$ meters traffic in its region bound for destination $d$ into the core; the self-destination coordinate is masked, leaving five effective controls.  If $x_{id}$ is the commodity occupancy and $n_i$ the region total, its core-entry rate is $u_{id}(x_{id}/n_i)G(n_i)$, with zero flow for an empty region.  The shared reward is completed trips per interval divided by 100.  The core constraint is $n_0\leq n_{\mathrm{safe}}$, where $n_{\mathrm{safe}}\simeq8786.415$ is the descending-branch solution of $G(n)=1$.

\section{Observations and Policy Network Architectures}
\label{sec:observation_policy_details}

This section describes the actor inputs and policy networks used in the reported benchmarks.

\subsection{Cooperative Transport}
\label{sec:transport_policy_details}

For agent $i$, the observation concatenates its absolute planar position and velocity (four coordinates), package position minus goal position (two), package position relative to the agent (two), absolute package velocity (two), and a binary package-on-goal indicator (one). It then includes the relative position of each of the $K$ obstacles ($2K$ coordinates) and the relative position and velocity of each of the other two agents (eight coordinates), ordered by agent index. Thus the observation dimension is $19+2K$, giving 21, 23, and 25 inputs for the one-, two-, and three-obstacle conditions. The nonrotating package contributes no orientation features, and no assigned push-slot features are appended. Actor inputs are standardized using running means and variances estimated during training and clipped to $[-10,10]$; evaluation uses the saved normalization statistics without updating them.

The actor is a feed-forward multilayer perceptron with two 64-unit hidden layers, tanh activations, and a linear two-dimensional Gaussian-mean head. A learned, observation-independent logstddev vector specifies the diagonal covariance. Training samples a latent Gaussian action and applies tanh to obtain the normalized force command in $[-1,1]^2$; deterministic evaluation applies tanh to the mean. The simulator multiplies this command by 0.6. MAPPO, SafeDiSTL, and the uniform-attribution variants share one actor across the three agents; MAPPO-L and MACPO use separate actor parameters for each agent with the same layer sizes and action parameterization.

\subsection{Connected Navigation}
\label{sec:navigation_policy_details}

Each agent receives a 72-dimensional observation. The first 34 coordinates comprise its own planar position and velocity (four), normalized episode time (one), four relative target positions (eight), and three teammate slots (21). The four target entries refer to the shared destination. Each teammate slot contains relative position and velocity from the stored communication packet, a direct-neighbor indicator, a connected-component reachability indicator, and packet age; slots follow the agents' fixed identifiers within the episode. The remaining 38 coordinates describe the ten obstacles using relative position and radius (30), the four corridor-boundary clearances (four), and the flattened two-dimensional world-rotation matrix (four). Positions, relative distances, obstacle radii, and boundary clearances are divided by 10; velocities are divided by the maximum speed of 1.5; time and packet ages are divided by the 250-step horizon. These features are then standardized using running observation statistics and clipped to $[-10,10]$, with the saved statistics held fixed during evaluation.

The actor maps these 72 inputs through two 64-unit tanh hidden layers to a linear two-dimensional Gaussian-mean head. It uses a learned, observation-independent diagonal log-standard-deviation vector and tanh-bounded actions in $[-1,1]^2$. Deterministic evaluation uses the transformed mean. The simulator additionally clips the command's Euclidean norm to one before applying the acceleration scale. MAPPO, SafeDiSTL, and the uniform-attribution variants share actor parameters across the four agents, whereas MAPPO-L and MACPO use one independently parameterized actor per agent with the same network dimensions. The policies are feed-forward; communication packet contents and ages are explicit observation features, not a recurrent policy state.

\subsection{Networked Traffic Control}
\label{sec:traffic_policy_details}

Each peripheral agent receives 31 observation coordinates: its six destination-specific queues, six local origin--destination demand rates, six destination-specific queues in the shared core, six current effective core-exit factors, normalized episode time, and a six-dimensional one-hot agent identity. Queue occupancies are divided by 1,800; demand rates and exit factors are used directly, and time is divided by the 240-step horizon. The core queues and exit factors are broadcast to all six agents. The observation excludes the latent variables driving future exit disturbances. The identity is already part of the environment observation, so the learner does not append another identity vector. There is no additional running observation normalization, and actor inputs use float32 arithmetic.

The actor has two 64-unit tanh hidden layers and a linear six-dimensional Gaussian-mean head, with a learned, observation-independent log-standard-deviation vector. A sigmoid transforms each latent Gaussian sample into an admission command in $[0,1]$. The self-destination coordinate is masked to zero, leaving five effective controls per agent; deterministic evaluation applies the same sigmoid and mask to the mean. MAPPO, SafeDiSTL, and the uniform-attribution variants share one actor across the six agents. MAPPO-L and MACPO instead maintain six independent actors with the same input features, hidden-layer sizes, and action transform.

\end{document}